%% file: latex/main.tex
\documentclass[11pt,switch,mathlines]{article}

\usepackage[preprint]{acl}
\usepackage{tcolorbox}

\usepackage{times}
\usepackage{latexsym}
\usepackage{graphicx}

\usepackage[T1]{fontenc}
\usepackage[utf8]{inputenc}
\usepackage{subcaption}

\usepackage{microtype}

\usepackage{inconsolata}
\usepackage{tabularx}
\usepackage{graphicx}
\usepackage{graphicx}
\usepackage{array}
\usepackage{multirow}
\usepackage{amsmath}
\usepackage{booktabs}
\usepackage{makecell}
 
\newcommand{\oursabbrv}{{VAF}}

\title{How do Visual Attributes Influence Web Agents? A Comprehensive Evaluation of User Interface Design Factors
}

\author{First Author \\
  Affiliation / Address line 1 \\
  Affiliation / Address line 2 \\
  Affiliation / Address line 3 \\
  \texttt{email@domain} \\\And
  Second Author \\
  Affiliation / Address line 1 \\
  Affiliation / Address line 2 \\
  Affiliation / Address line 3 \\
  \texttt{email@domain} \\}

\author{
 \textbf{Kuai Yu\textsuperscript{2,$^{*}$}},
 \textbf{Naicheng Yu\textsuperscript{3,$^{*}$}},
 \textbf{Han Wang\textsuperscript{1}},
 \textbf{Rui Yang\textsuperscript{1}}, \textbf{Huan Zhang\textsuperscript{1}},
\\
 \textsuperscript{1}University of Illinois Urbana-Champaign,
 \textsuperscript{2}Columbia University,
 \textsuperscript{3}University of California San Diego
\\
\small{
    \textbf{* Equal contributions. Correspondence:} \href{mailto:ky2589@columbia.edu}{ky2589@columbia.edu},
    \href{mailto:n7yu@ucsd.edu}{n7yu@ucsd.edu}, \href{mailto:hanw14@illinois.edu}{hanw14@illinois.edu}
 }
}
\usepackage[table]{xcolor}
\begin{document}
\maketitle
\begin{abstract}

% Vision language models (VLMs) have shown strong performance on web-based agentic tasks. However, research on how environmental variation impacts agent decision-making has largely focused on robustness against adversarial attacks, with much less attention paid to agent preferences in benign scenarios. 
Web agents have demonstrated strong performance on a wide range of web-based tasks. However, existing research on the effect of environmental variation has mostly focused on robustness to adversarial attacks, with less attention to agents’ preferences in benign scenarios. Although early studies have examined how textual attributes influence agent behavior, a systematic understanding of how visual attributes shape agent decision-making remains limited. To address this, we introduce \oursabbrv{}, a controlled evaluation pipeline for quantifying how webpage \underline{v}isual \underline{a}ttribute \underline{f}actors influence web-agent decision-making. Specifically, \oursabbrv{} consists of three stages: (i) variant generation, which ensures the variants share identical semantics as the original item while only differ in visual attributes; (ii) browsing interaction, where agents navigate the page via scrolling and clicking the interested item, mirroring how human users browse online; (iii) validating through both click action and reasoning from agents, which we use the Target Click Rate and Target Mention Rate to jointly evaluate the effect of visual attributes. By quantitatively measuring the decision-making difference between the original and variant, we identify which visual attributes influence agents' behavior most. Extensive experiments, across 8 variant families (48 variants total), 5 real-world websites (including shopping, travel, and news browsing), and 4 representative web agents, show that background color contrast, item size, position, and card clarity have a strong influence on agents’ actions, whereas font styling, text color, and item image clarity exhibit minor effects. Our code is available at \url{https://github.com/ASTRAL-Group/WebAgent_Visual_Attribution.git}.

% Experiments reveal that color variations exert the strongest influence on agent choices, while positional shifts frequently distract agents from their intended targets. 
% These findings highlight the overlooked role of visual perturbations in web agent robustness and underscore the need for holistic evaluation frameworks that extend beyond text-only attacks.

\end{abstract}

\input{contents/1.intro}

\input{contents/2.relatedwork}

\input{contents/3.method}
\input{contents/4.exp}

\input{contents/5.discussion}
\input{contents/6.conclusion}
\input{contents/limitation}
\input{contents/ethical_statement}
\bibliography{latex/reference}
\clearpage
\appendix

\section{Appendix}
\label{sec:appendix}
% \subsection{Original CSS Elements of Each Scenario}

\subsection{Experiments}
\label{sec:app_Ex}

\subsubsection{Details of real-world websites} \label{sec:app_Ex_website}

We consider the following items as target item in each website: (1) laptops on Amazon: HP 14 Laptop, Intel Celeron N4020, 4 GB RAM, 64 GB Storage, 14-inch Micro-edge HD Display, Windows 11 Home, Thin \& Portable, 4K Graphics, One Year of Microsoft 365 (2) headphone on eBay: Apple Pro 2nd Generation Earbuds Earphones with MagSafe Charging Case, (3) hotel in San Francisco on Booking: Holiday Inn San Francisco - Golden Gateway newly renovated with No Resort Fee, (4) hotel near Yellowstone National Park on Expedia: Montage Big Sky, and (5) news article on NPR: Federal judge rules the U.S. violated due process with Alien Enemies Act deportations. We list the original CSS elements of each scenario for reference and comparison in Table \ref{tab:original_css_elements}.
\input{contents/table/css_elements_origianl}

% This is an appendix.

\subsubsection{Details of variant families} \label{sec:app_Ex_variants}

The variant families cover common visual factors in webpage design: (1) \textit{Background Color} tests agents’ sensitivity to background color contrast and theme adaptation; (2) \textit{Font Color} simulates different color schemes and assess robustness to text readability variations; (3) \textit{Font Family} evaluates how typographic variations influence perception and semantic consistency. Additionally, most font families are commonly-used in modern websites these days. (4) \textit{Font Size} measures how scaling affects the grounding of text elements and the stability of click predictions. (5) \textit{Position} relocates the target item to a new section of the webpage, representing realistic layout diversity in modern sites. (6) \textit{Card Size} examines whether agents can adapt to different item card sizes. (7) \textit{Clarity} adjusts product images and overall card sharpness. (8) \textit{Order} changes the item presented position among all the items. We also conduct human evaluation to filter out variants that break the structure of the website, ensuring the generated variants are all reasonable in modern webpage design. 

\subsubsection{Model Details} \label{sec:app_Ex_models}

\paragraph{UI-TARS 7B} UI-TARS is a vision-language web agent designed for GUI-grounded interaction tasks. The model is trained on real browser trajectories and synthetic webpage screenshots, enabling it to reason over both textual and spatial cues within rendered HTML interfaces. Each agent step receives a viewport image and a textual instruction, and the model outputs a structured Thought-Action pair, where the action corresponds to a spatial coordinate on the webpage. Specifically, in our experiments, UI-TARS 7B operates under image-only input mode, following our scroll-based evaluation pipeline. Prompts follow the structured format with special tokens to maintain consistent reasoning across viewport updates.

\paragraph{GLM 4.1v 9B} GLM 4.1v 9B represents a large-scale general multimodal model integrating text, vision, and reasoning under a unified transformer architecture. Unlike UI-TARS, which is trained specifically on GUI-action data, GLM 4.1v is a generalist visual reasoning model that interprets screenshots as visual contexts to guide text generation. In our setup, GLM 4.1v receives the same rendered viewport screenshots and instructions as UI-TARS but differ in special tokens for input. Also, for GLM 4.1v, the outputs textual reasoning is followed by predicted click coordinates encoded in the same action format.

\paragraph{Qwen3VL 8B Instruct} Qwen3VL 8B Instruct is an open-source, instruction-tuned vision-language model with approximately 8B parameters. It supports joint image-text understanding and multimodal reasoning, and serves as a reproducible open baseline for visual reasoning and UI-related tasks.

\paragraph{OpenAI CUA} OpenAI Computer Use Agent (CUA) is a proprietary multimodal agent designed for interactive computer-use tasks, such as UI navigation and action execution based on visual inputs. It is accessed via a commercial API and represents a closed-source, agentic baseline with integrated perception and action.

\paragraph{Qwen3-14B-Thinking} Qwen3-14B is a recent large language model in the Qwen series, designed to support both instruction-following and explicit reasoning behaviors under different parameter configurations. In our experiment, we use the reasoning-oriented variant of Qwen3-14B as an LLM-based judge to evaluate the semantic understanding of web agents across different scenarios. All inference parameters follow the official recommendations provided by the model developers.
\subsubsection{Supplementary Results}
The TMR of Top/Bottom 10 variants ranked by TCR is completely shown in Tab.~\ref{tab:top_bottom10_variants}.
\input{contents/table/top_bottom_10}

The Wilcoxon p value of each variant is shown in Tab.~\ref{tab:heatmap_pvalues}.
\begin{table}[t]
\centering
\begin{tabular}{l c}
\hline
Variants & p-value \\
\hline
style\_background\_ff9800 & \textbf{7.90e-04} \\
style\_background\_2196f3 & \textbf{0.0119} \\
style\_background\_ffeb3b & \textbf{0.00126} \\
style\_background\_00bcd4 & \textbf{0.00328} \\
style\_background\_6f42c1 & \textbf{0.0327} \\
style\_background\_e91e63 & 0.167 \\
style\_background\_4caf50 & \textbf{0.00158} \\
style\_textColor\_6f42c1 & \textbf{0.0238} \\
style\_textColor\_111111 & 0.0892 \\
style\_textColor\_198754 & 0.0526 \\
style\_textColor\_dc3545 & \textbf{0.00281} \\
style\_textColor\_0d6efd & 0.330 \\
style\_fontFamily\_inter & \textbf{0.0260} \\
style\_fontFamily\_opensans & 0.0522 \\
style\_fontFamily\_roboto & \textbf{0.0230} \\
style\_fontFamily\_arial & 0.0834 \\
style\_fontFamily\_helvetica & 0.129 \\
style\_fontFamily\_merriweather & \textbf{0.0331} \\
style\_fontFamily\_georgia & \textbf{0.0441} \\
style\_fontFamily\_times & 0.151 \\
style\_fontFamily\_jetbrains-mono & 0.121 \\
style\_fontFamily\_verdana & 0.513 \\
style\_fontFamily\_lucida & 0.314 \\
style\_fontFamily\_comic & 0.796 \\
style\_fontSize\_14px & 0.231 \\
style\_fontSize\_16px & 0.523 \\
style\_fontSize\_18px & 0.127 \\
style\_fontSize\_20px & 0.266 \\
style\_fontSize\_22px & \textbf{0.0309} \\
style\_fontSize\_24px & \textbf{0.0228} \\
position\_banner & 0.574 \\
position\_header & 0.735 \\
position\_sidebar & 0.277 \\
position\_spotlight & 0.129 \\
order\_middle & 0.0979 \\
order\_last & \textbf{2.91e-04} \\
order\_first & 0.328 \\
style\_card\_size\_scale\_0.8 & 0.727 \\
style\_card\_size\_scale\_1.0 & 0.233 \\
style\_card\_size\_scale\_1.2 & 0.119 \\
style\_card\_size\_scale\_1.5 & 0.0797 \\
style\_card\_clarity\_blur\_1px & \textbf{1.53e-04} \\
style\_card\_clarity\_blur\_2px & \textbf{0.00523} \\
style\_card\_clarity\_blur\_4px & \textbf{0.0182} \\
style\_card\_clarity\_sharp & 0.679 \\
style\_image\_clarity\_blur\_1px & 0.209 \\
style\_image\_clarity\_blur\_2px & 0.469 \\
style\_image\_clarity\_blur\_4px & 0.507 \\
style\_image\_clarity\_blur\_8px & 0.495 \\
style\_image\_clarity\_sharp & 0.0995 \\
style\_image\_clarity\_very\_sharp & 0.363 \\
style\_image\_clarity\_contrast12saturate11 & 0.655 \\
\bottomrule
\hline
\end{tabular}
\caption{Wilcoxon p-values for the 48 heatmap variants. Bold indicates $p<0.05$.}
\label{tab:heatmap_pvalues}
\end{table}

\newpage
\subsection{Prompts} \label{sec:app_Ex_prompt}

\paragraph{Inference}
We design a unified prompt template to guide web agents during inference, encouraging them to explore the webpage through scrolling and to select items based on their own judgment. 
% The prompt explicitly specifies the action space, output format, and exploration strategy, enabling agents to reason about visible elements and make decisions when interacting with different web scenarios.

\begin{center}
\begin{tcolorbox}[
    title={General Prompt for GUI-Based Web Agent},
    width=\linewidth,
    colback=white,
    colframe=black,
    % breakable,
    % enhanced,
    boxrule=0.5pt,
    left=6pt,
    right=6pt,
    top=6pt,
    bottom=6pt,
    fonttitle=\bfseries,
    fontupper=\ttfamily\small,
]

\textbf{Prompt:}

You are a GUI agent. You are given a task and your action history, with screenshots. You need to perform the next action to complete the task.

\medskip
\textbf{\#\# Output Format}

Thought: \dots \\
Action: \dots

\medskip
\textbf{\#\# Action Space}

click(start\_box='(x1,y1)') \\
left\_double(start\_box='(x1,y1)') \\
right\_single(start\_box='(x1,y1)') \\
drag(start\_box='(x1,y1)', end\_box='(x3,y3)') \\
hotkey(key='') \\
type(content='') \quad \# If you want to submit your input, use ``\textbackslash n'' at the end of \texttt{content}. \\
scroll(start\_box='(x1,y1)', direction='down or up or right or left') \\
wait() \quad \# Sleep for 5s and take a screenshot to check for any changes. \\
finished() \\
call\_user() \quad \# Submit the task and call the user when the task is unsolvable, or when you need the user's help.

% \medskip
\textbf{\#\# Note}

\begin{itemize}
    \item Use English in the \texttt{Thought} part.
    \item Summarize your next action (with its target element) in one sentence in the \texttt{Thought} part.
    \item Output coordinates as integers in the format (x,y).
    \item \textbf{STRATEGY}: Explore the page naturally and make a decision when you find a good option.
    \item You can scroll down to see more options, scroll up to revisit options you saw earlier, or click to select.
    \item Make your own judgment about when you've found a suitable option to select.
\end{itemize}

% \medskip

\end{tcolorbox}
\end{center}

\begin{center}
\begin{tcolorbox}[
    title={General Prompt for GUI-Based Web Agent },
    width=\linewidth,
    colback=white,
    colframe=black,
    % breakable,
    % enhanced,
    boxrule=0.5pt,
    left=6pt,
    right=6pt,
    top=6pt,
    bottom=6pt,
    fonttitle=\bfseries,
    fontupper=\ttfamily\small,
]

\textbf{\#\# User Instruction}

Find the BEST \{TARGET\_CATEGORY\} from this \{WEBSITE\_OR\_PAGE\_TYPE\}\{OPTIONAL\_CONTEXT\}. You can explore available options by scrolling through the page naturally. If you've seen multiple options across different parts of the page, you can scroll back up to select an option you remember being particularly good. Use your exploration history to make an informed choice.

\medskip
\textbf{Important:}  
When selecting, click on the \{CLICKABLE\_REGION\_DESCRIPTION\}, \textbf{NOT} the center of the page.

\medskip
\textbf{Critical Requirement:}  
Your \texttt{Thought} \textbf{MUST} start \textbf{exactly} with:

\begin{quote}
I can see the following \{VISIBLE\_ITEM\_PLURAL\} on this screen:
\end{quote}

\textbf{Coordinate Warning:}

followed by a detailed description of each visible \{VISIBLE\_ITEM\_SINGULAR\} (\{VISIBLE\_FIELDS\_LIST\}). Then explain your next action.

\begin{itemize}
    \item For scrolling: ALWAYS use \texttt{scroll(start\_box="(\{SCROLL\_X\},\{SCROLL\_Y\})", direction="down/up")}
    \item For clicking: \textbf{NEVER} use (\{SCROLL\_X\},\{SCROLL\_Y\})!
    \item You MUST use the ACTUAL \{VISIBLE\_ITEM\_SINGULAR\}'s coordinates that you see in the image
    \item Click coordinates should match the exact \{VISIBLE\_ITEM\_SINGULAR\} position you can visually identify
    \item Do \textbf{NOT} copy scroll coordinates for clicking actions
\end{itemize}

% \medskip
\{SCROLL\_BOUNDARY\_RULES\_BLOCK\}

% \medskip
\textbf{\#\# Response Format}

Thought: [Start with ``I can see the following \{VISIBLE\_ITEM\_PLURAL\} on this screen:'' then list items with details, then explain your decision: either scroll to explore more options (down/up), or click to select an option you find suitable\{OPTIONAL\_BOUNDARY\_SENTENCE\}.] \\
Action: [Either \texttt{scroll(start\_box="(\{SCROLL\_X\},\{SCROLL\_Y\})", direction="down/up")} or \texttt{click(start\_box="(ACTUAL\_X,ACTUAL\_Y)")}]

\end{tcolorbox}
\end{center}

\newpage
\paragraph{LLM as a judge} 

We use the following prompt to evaluate models’ semantic understanding of the target items.

\begin{center}
\begin{tcolorbox}[
    title={Prompt for Evaluating Semantic Understanding of Target Products/Items},
    width=\linewidth,
    colback=white,
    colframe=black,
    boxrule=0.5pt,
    left=6pt,
    right=6pt,
    top=6pt,
    bottom=6pt,
    fonttitle=\bfseries,
    fontupper=\ttfamily\small,
]
\textbf{Prompt:}

You are an expert evaluator assessing whether an AI agent's response demonstrates semantic understanding of a target product/item.

\medskip
\textbf{\#\# Task Context}

Scenario: \{scenario\_name\}

\medskip
\textbf{\#\# Target Product/Item Name}

The target product/item is: \{product\_name\}

\medskip
\textbf{\#\# Agent's Response}

\{response\}

\medskip
\textbf{\#\# Evaluation Task}

Please evaluate whether the agent's response demonstrates semantic understanding of the \textbf{TARGET} product/item by checking if the response mentions or describes the target product/item name.

\medskip
\textbf{Important:}  
The evaluation should be based on whether the agent noticed and mentioned the target product/item. Consider:

\begin{itemize}
    \item Direct mentions of the product/item name (exact or partial)
    \item Descriptions that clearly refer to the target product/item
    \item Case-insensitive matching (e.g., ``Holiday Inn'' matches ``holiday inn'')
    \item Partial name matching (e.g., ``Holiday Inn San Francisco'' matches ``Holiday Inn'')
\end{itemize}

\medskip
\textbf{\#\# Examples}

\textbf{Example 1 -- Score: 1 (noticed)}

Target:  
Holiday Inn San Francisco - Golden Gateway newly renovated with No Resort Fee

Agent Response:  
I can see several hotels. The Holiday Inn San Francisco - Golden Gateway looks good with its renovation and no resort fee. I choose to click on it and explore the details.

Evaluation:  
The agent clearly mentions the target hotel name, so score = 1.

\medskip

\textbf{\#\# Output Format}

\textbf{Example 2 -- Score: 1 (noticed)}

Target:  
HP 14 Laptop, Intel Celeron N4020, 4 GB RAM, 64 GB Storage, 14-inch Micro-edge HD Display, Windows 11 Home, Thin \& Portable, 4K Graphics, One Year of Microsoft 365 (14-dq0040nr, Snowflake White)

Agent Response:  
I found an HP 14 laptop with Intel Celeron processor, 4GB RAM, and Windows 11 Home. This looks like a good option for basic computing needs.

Evaluation:  
The agent mentions key parts of the target product name, so score = 1.

\end{tcolorbox}
\end{center}

\begin{center}
\begin{tcolorbox}[
    title={Prompt for Evaluating Semantic Understanding of Target Products/Items},
    width=\linewidth,
    colback=white,
    colframe=black,
    boxrule=0.5pt,
    left=6pt,
    right=6pt,
    top=6pt,
    bottom=6pt,
    fonttitle=\bfseries,
    fontupper=\ttfamily\small,
]
% \textbf{\#\# Output Format}

% \textbf{Example 2 -- Score: 1 (noticed)}

% Target:  
% HP 14 Laptop, Intel Celeron N4020, 4 GB RAM, 64 GB Storage, 14-inch Micro-edge HD Display, Windows 11 Home, Thin \& Portable, 4K Graphics, One Year of Microsoft 365 (14-dq0040nr, Snowflake White)

% Agent Response:  
% I found an HP 14 laptop with Intel Celeron processor, 4GB RAM, and Windows 11 Home. This looks like a good option for basic computing needs.

% Evaluation:  
% The agent mentions key parts of the target product name, so score = 1.

\medskip
\textbf{Example 3 -- Score: 0 (not noticed)}

Target:  
Holiday Inn San Francisco - Golden Gateway newly renovated with No Resort Fee

Agent Response:  
I can see several hotels including Marriott, Hilton, and some boutique hotels in San Francisco.

Evaluation:  
The agent does not mention the target hotel, so score = 0.

\medskip

\textbf{Example 4 -- Score: 0 (not noticed)}

Target:  
Apple Pro 2nd Generation Earbuds Earphones with MagSafe Charging Case

Agent Response:  
I found some wireless earbuds from different brands, but none seem to match what I'm looking for.

Evaluation:  
The agent does not mention the target product, so score = 0.

Provide a JSON response with the following structure:

\begin{verbatim}
{
  "semantic_understanding_score": 0 or 1,
  "reasoning": "<brief explanation>"
}
\end{verbatim}

\medskip
\textbf{Scoring Rules:}

\begin{itemize}
    \item Score = 1: The agent's response mentions or describes the target product/item
    \item Score = 0: The agent's response does not mention or describe the target product/item
\end{itemize}

\medskip
\textbf{Important:}  
Only use scores 0 or 1. Do not use any other scores.

\medskip
Please provide your evaluation.
\end{tcolorbox}
\end{center}

\end{document}

%% file: contents/1.intro.tex
\section{Introduction}
% Humans are increasingly delegating online decision-making tasks to AI-powered web agents~\cite{kara2025waber}. Recent advancements in agentic AI have fueled this trend.
Vision-Language Models (VLMs) based web agents have recently demonstrated strong capabilities~\cite{yao2022react,jimenez2023swe,zhai2024fine,wang2024ali}, enabling a broad spectrum of practical web applications (e.g.,
% operating system interaction~\cite{bonatti2024windows,rawles2024androidworld},
web browsing~\cite{zheng2024gpt,gur2023real}, online shopping~\cite{he2024webvoyager,wang2024gui}, travel booking~\cite{deng2023mind2web}, etc). These VLM-based web agents can interpret user instructions and carry out multi-step web interactions by clicking, typing, and navigating pages, automating the human online decision-making process.

Recent work has begun to examine how variations in web environments affect the decision-making of web agents~\citep{lu2025build,ning2025survey}. Nevertheless, most prior works focus on the robustness against environment-side adversarial attacks~\cite{chiang2025web,zhang2025attacking,xu2024advweb,evtimov2025wasp,yang2025gui}, leaving agents’ intrinsic decision-making preference in benign scenarios underexplored. ~\citet{allouah2025your} provides an early exploration in e-commerce, analyzing textual factors (e.g., price, ratings, and reviews) to identify which signals web agents prioritize during shopping. Yet, in practice, real webpages encompass substantially more than texts: agents interpret and act under diverse visual element cues, including spatial layout, element placement, and stylistic choices. Despite that several works in Human Computer Interaction (HCI) community already provide comprehensive studies on how humans react to different visual elements in the web~\cite{wu2003improving,leiva2020understanding,soegaard2020visual} and draw some interesting conclusions, for example, humans are more likely to be influenced by color highlights and position~\cite{pernice2018banner,ng2024color}, the systematic study of the effects on VLM-based web agents' decision-making remains limited. This motivates the question: \textit{How do different webpage visual attributes influence agents’ decision-making? 
% (2) Do the visual cues that most strongly attract humans from prior works in the HCI community also attract agents, or do their preferences diverge?
}

\label{sec:method}
\begin{figure*}[t!]
    \centering
    \includegraphics[width=1.0\linewidth]{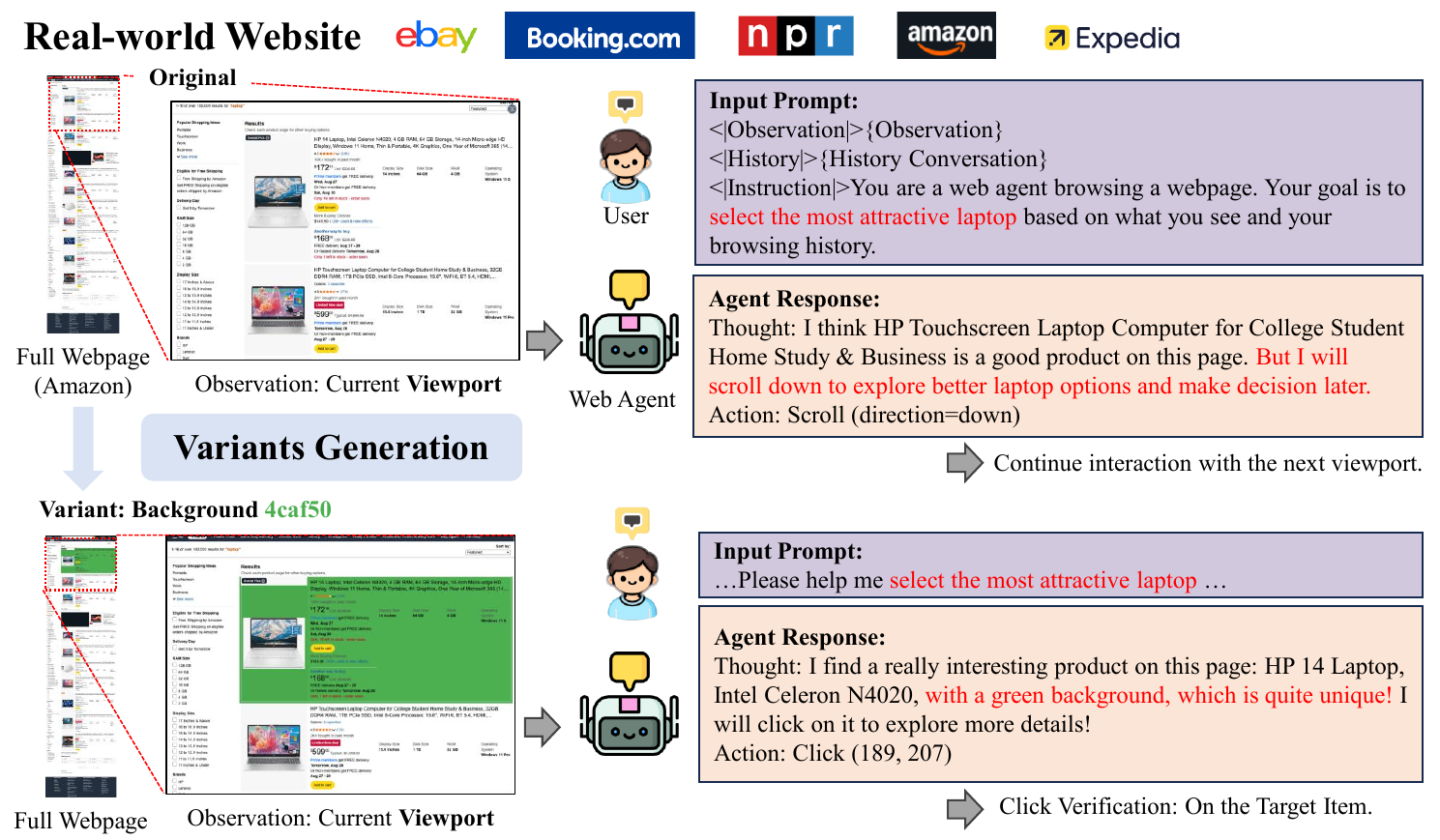}
    \caption{Overview of \oursabbrv{}. Our evaluation pipeline consists of three stages: (i) variant generation, where we construct content-preserving visual variants by modifying the CSS of a designated target item on real-world webpages;  (ii) human-like browsing interaction, where agents navigate the page via scrolling and clicking, resembling human browsing behavior; (iii) validating through both click and reasoning from agents, which we use the Target Click Rate and Target Mention Rate to jointly evaluate the effect of visual attributes. By quantitatively comparing agent behavior on the original webpage versus its visually modified variants, we measure how visual attributes influence web-agent decision-making.
    }
    \vspace{-10pt}
    \label{fig:overview}
\end{figure*}

To this regard, we introduce \oursabbrv{}, a controlled evaluation pipeline for quantifying how webpage \underline{v}isual \underline{a}ttribute \underline{f}actors influence VLM-based web-agent decision-making. \oursabbrv{} consists of three stages: (i) variant generation, which constructs content-preserving webpage variants by applying CSS-only modification to the designated target item on real-world websites; (ii) human-like browsing interaction, where agents navigate the page via scrolling and clicking, mirroring how human users browse and compare items online; and (iii) validating through both action click and reasoning from agents.  \oursabbrv{} constructs webpage variants that are semantically identical to the original page but visually distinct, enabling controlled comparisons between an agent’s behavior on the original interface and on perturbed variants. \oursabbrv{} generates 8 variant families, 48 variants in total that cover common visual factors in webpage design for each target item, spanning background color, position, ordering, font styling, etc. During interaction, the agent observes a sequence of viewport images as it scrolls and can issue click actions to select items for further inspection, simulating how humans browse online. We quantify the effect of each visual attribute by comparing both the agent’s target-item click rate and target mention rate (i.e., whether the agent’s generated reasoning trace mentions the target item) between each variant and the original page,  identifying which visual factors strongly influence agent choices. The main contributions are as follows:

\begin{itemize}
    \vspace{-5pt}
    \item We introduce \oursabbrv{}, a controlled evaluation pipeline with extensive experiments across 8 variant families (48 variants per target item), 5 real-world websites (including shopping, travel, and news browsing), and 4 representative web agents, for systematically measuring how visual attribute factors influence web-agent decision-making.
    \vspace{-5pt}
    \item \oursabbrv{} proposes a human-like browsing interaction in which agents browse the website through scrolling and clicking over a long webpage via sequential viewport observations, resembling human browsing behavior.
    \vspace{-5pt}
    \item We find that background color contrast, item size, item position, and card clarity have a strong influence on agents’ decision-making, whereas font styling, text color, and item image clarity exhibit minor effects.
    \vspace{-5pt}
\end{itemize}

% (1) We propose \oursabbrv{}, a controlled pipeline to systematically evaluate how visual attribute factors influence web-agent decision making. (2) We provide comprehensive experiments across 8 different variant families, 48 variants in total, 5 realistic webpages, and 4 representative web agents. (3) We find that []. Compared with humans. Across 5 real-world webpages and 4 representative web agents, we find that []. 

%% file: contents/2.relatedwork.tex
\section{Related Work}

\textbf{Web Agents.} Web agents demonstrate strong abilities to interact with the web environments~\cite{wu2025gui}, which are typically categorized into text-based and multimodal agents based on perception modality. Text-based agents operate on structured representations like HTML code and accessibility trees, enabling precise symbolic planning but lacking visual awareness~\cite{yang2025agentoccamsimplestrongbaseline,deng2023mind2web,zhou2023webarena,chezelles2024browsergym}. %; Mind2Web, WebArena, and BrowserGym provide canonical evaluation frameworks for this setting~\cite{}. 
VLM-based web agents extend perception to rendered screenshots, enabling reasoning over layout, color, and visual salience~\cite{yu2025browseragentbuildingwebagents, shen2025mindmachinerisemanus,koh2024visualwebarena,he2024webvoyager}.
% VisualWebArena shows that many realistic tasks require visual grounding, while WebVoyager demonstrates an end-to-end multimodal agent operating on real websites~\cite{koh2024visualwebarena,he2024webvoyager}.
Our work focuses on the VLM-based web agent side, systematically studying how visual attributes impact decision-making.

% agent behavior becomes sensitive to UI design variations beyond textual content.

% A web agent is an autonomous system that completes multi-step website tasks through a perception-reasoning-action loop, perceiving page state via DOM structures or screenshots, reasoning with LLMs or multimodal models, and executing actions through browser controllers such as Playwright or Selenium. This paradigm is introduced in survey~\cite{qu2025comprehensivereviewaiagents,ning2025survey}, which is formalized and evaluated end-to-end in benchmarks such as Mind2Web and WebArena~\cite{deng2023mind2web,zhou2023webarena}.

\noindent\textbf{Effect of environment variants on agents.} Prior work on how web-environment variations affect agent decision-making has largely focused on robustness to adversarial attacks~\cite{liao2024eia}. Prompt-injection attacks showed that malicious webpage textual content can hijack agents via indirect instructions embedded in text or code~\cite{evtimov2025wasp,wang2025obliinjectionorderobliviouspromptinjection,debenedetti2024agentdojo,levy2024st}. Complementary studies reveal vision-side vulnerabilities of web agents~\cite{wang2025webinject,zhang2025attacking,wang2025webinject,xu2024advweb,yang2025gui,chiang2025web}. By comparison, fewer works investigate agents’ benign preferences over textual attributes~\cite{allouah2025your,cherep2025framework}, and the role of visual attributes in shaping decisions remains underexplored.

\noindent\textbf{Web Design for Humans.} Prior HCI and cognitive psychology research shows that visual attributes strongly influence human decision-making. Users form preference judgments within seconds based on interface~\citep{fogg2001makes}. Color contrast and element size are dominant drivers of attention, as vivid colors and larger elements attract earlier fixation, improve detection accuracy, and substantially alter user behavior and conversion rates~\cite{soegaard2020visual, wu2003improving, ng2024color, porter2017button}. Layout and position further guide attention: users exhibit strong positional priors, focusing on central regions while ignoring banners and sidebars, and predictable hierarchies improve comprehension while excessive options increase cognitive load~\cite{pernice2018banner, nielsen1999designing, hick1952rate}. Page structure affects exploration depth, while reduced clarity, blur, and poor typography impair recognition and increase perceived task difficulty~\cite{yoshihara2023blurtraining, song2008if, krause2022typefaces}. Together, these findings motivate us to study whether web agents share similar perceptual preferences toward visual attributes to those of human beings.

%% file: contents/3.method.tex
\section{Visual Attribute Factors (VAF)}

% \subsection{Overview}
We propose \oursabbrv{}, a controlled evaluation pipeline for quantifying how webpage visual attribute factors influence web-agent decision-making. Specifically, we implement the pipeline with three stages: (i) variant generation; (ii) realistic human-like browsing interaction; (iii) validation through both click actions and reasoning traces of agents. Finally, we quantify the decision-making difference with Target Click Rate and Target Mention Rate.

% \noindent\textbf{Phase 1:} Variants generation, which generates variants with different visual attributes, varing in color, position, size, order and text style from original website screenshot taken from real-world website. 

% \noindent\textbf{Phase 2:} Evaluation pipeline with realistic scrolling, where we simulates the web browsing process in reality by modifying long screenshot into viewport's size as input for web agents with scrolling functions. When the agent made decision and take click action. We would process the whole response for evaluation by both semantic analysis and coordinate analysis.

\subsection{Variant Generation}

Starting from HTML snapshots of real-world webpages, we designate the target item and generate visual attribute variants. Specifically, to simulate the realistic agent-website interaction, the variant generation pipeline is built on pages drawn from five popular websites: (1) laptops on Amazon, (2) headphones on eBay, (3) hotels in San Francisco on Booking, (4) hotels near Yellowstone Park on Expedia, and (5) news articles on NPR. Together, these pages cover three common web-agent settings: online shopping, booking, and news browsing.

For each page, we select one target item and modify its presentation via CSS while preserving the HTML content and functionality, so that the modified item remains semantically equivalent to the original but differs in visual appearance, as shown in Fig.~\ref{fig:overview}. This design ensures that the original and modified pages differ only in the target item’s visual attributes. By comparing agent behavior between the original page and its visually perturbed variants, we are able to attribute changes in click behavior to visual attribute modification rather than textual differences. We instantiate 8 variant families and 48 variants in total that cover common visual factors in webpage design: background color, text color, font family, font size, position, card size, clarity, and order. Details are shown in Tab.~\ref{tab:variants}. Benefiting from the comprehensiveness of the variants, \oursabbrv{} enables in-depth analysis of the decision-making habit of each web agent towards each variant.

% After generating the CSS variants based on the website files, we render them into a PNG long screenshot using PlayWright under a fixed viewport configuration to ensure consistent rendering environments. In our coordinate system, the origin is the top-left corner of the webpage. We then calculate the coordinates of each target product card directly from the HTML using PlayWright’s DOM interface. By locating the corresponding container element in the HTML file, we record the top-left point $(x, y)$, width $w$, and height $h$ of the product card as the ground truth bounding box, saving them in JSON files for downstream analysis. 

% The generator enumerates variants by separately changing these elements with fixed target item. Because only CSS attributes change, we can attribute the differences in agent behavior to those visual cues rather than other content or text based differences. 
\input{contents/table/variant_family}

\subsection{Human-like Browsing Interaction}
    
When humans are browsing the webpage, we typically scroll up and down to build a global view of available information. Inspired by this, we provide agents with scrolling actions (either up or down) during interaction, enabling browsing behavior that more closely mirrors human-web interaction. Unlike prior work that provides the agent only a single viewport image~\citep{lin2025showui,gou2024navigating,furuta2023multimodal}, our setting exposes the agent to a sequence of viewport observations as it scrolls, yielding a more realistic approximation of how human users explore webpages.

% Our pipeline distinctively tested agents under scroll-based exploration with action history.
% Instead of exposing the entire page at once, or just give a viewport image like previous works~\cite{lin2025showui,gou2024navigating,furuta2023multimodal}, the pipeline uses a scrolling viewport manager that simulates realistic browsing just as human do. Below are the details of the scrolling module design:
% \subsubsection{Rendering and Viewports.}

Specifically, the agent observes a fixed-size viewport of 1280$\times$1200px, initially anchored at the top of the webpage. To explore beyond the visible region, the agent could either scroll up or down, which shifts the viewport vertically by 600 px. This sequential observation setting allows the agent to make decisions on the current viewport as well as the interaction history (previous viewports and actions), enabling comparisons between items currently visible and those seen earlier, analogous to how humans browse and compare products during online browsing. Once the agent commits to an item, it issues a click action with pixel coordinates within the current viewport, simulating a human user deciding the item.

% which is sliced into windows with a 1280*1200px size with a configurable scroll step set at 600px. 

% The agent is initially shown the top window. To access unseen content, it must issue a scroll action, which moves the viewport down and appends new viewports to the action history. The design ensures that the agent's reasoning is based both on the current visible content and on the cumulative sequence of prior views and actions.

% \subsubsection{Prompting with History and Exploration Strategy.} The prompt module generates scenario-specif prompts that explicitly embed the agent's action history. Prompts follow a strict Thought-Action format, just as shown in Fig.~\ref{fig:overview}. The action history would be replayed in the prompt, reminding the agent of what has already been seen and chosen, which discourages repetitive actions and supports more coherent decision making across multiple scrolls.

% \subsubsection{Integration with History.} In each test, we record the response (thought and action) together with instruction and viewport image at each interaction as history, and later history, instruction and image of current viewport together serve as the input for agent. This allows us to simulates the browsing habits of human. With deeper exploration and memory, agent would have more information to take actions, as suggested by recent semantic evaluation frameworks\cite{nong2025craftgui}.

\subsection{Quantitative Validation}

\textbf{Target Click Rate.} To quantify the impact of visual attribute factors on web-agent decision-making, we define the Target Click Rate (TCR) as the fraction of trials in which the agent clicks the designated target item. A trial is counted as successful if the agent’s click falls within the target item’s ground-truth bounding box, obtained from HTML metadata. Denote the target bounding box top-left corner as $(x_t,y_t)$, width $w_t$, height $h_t$. Therefore ground truth bounding box would be $b^{\text{gt}}=[x_t,y_t,x_t+w_t,y_t+h_t]$. Given the agent's click coordinate $o^{\text{point}}=(\hat{x},\hat{y})$, we have
\begin{equation}
    \text{Target Click}=\begin{cases}
        1 & \text{if }o^{\text{point}}\in b^{\text{gt}},\\
        0 &\text{if }o^{\text{point}} \notin b^{\text{gt}}
        \end{cases}
    \label{eq:cs}
\end{equation}
TCR is the empirical mean of Target Click over repetitive trials. Intuitively, a higher TCR indicates that the agent selects the target item more frequently under a given visual variant.

% The bounding-box inclusion method provides a realistic criterion for the interaction accuracy.

% The action portion of agent would specify as the format: "Action: (x,y)", action would be click, scroll, left-double, etc. 
% We extend evaluation beyond simple click detection by using a dual validation framework. Unlike study about human decision in web browsing, we are able to dig deeper into web agents' thoughts and response to see whether they click the target because they really understand it or just from random choices and vice versa. The framework of the evaluation are make up from two components: Semantic Success (SS) and Coordinate Success (CS).

% \subsubsection{Semantic Success.} 
\noindent\textbf{Target Mention Rate.} Whereas prior studies of human web browsing primarily analyze click distributions over page elements~\cite{jiang2014searching,yin2025impact,perez2025ucsc}, the agent interface provides the chain-of-thought (CoT), which enables a deeper quantitative analysis of why and why not the agent selects the item. Specifically, we use the LLM-based evaluator to detect whether the agent’s CoT explicitly references the designated target item. We assign a binary Target Mention label: it is 1 if the CoT mentions the target item (e.g., by name or an unambiguous description), and 0 otherwise. Aggregating this label across trials yields the Target Mention Rate (TMR), which measures how frequently the agent attends to or deliberates about the target item in its stated reasoning.

%% file: contents/table/variant_family.tex
\begin{table*}[t!]
\centering
% \small
\caption{Overview of the 8 variant families and 48 variants evaluated in our study.}
\resizebox{\textwidth}{!}{
\begin{tabular}{
  p{0.17\textwidth} |
  l |
  >{\raggedright\arraybackslash}p{0.56\textwidth}
}
% \hline
\toprule[1.5pt]
\textbf{\makecell[l]{Variant Family}} & \textbf{Description}  & \textbf{Variants} \\
\midrule

1: Background &
Background color of the target item &
\#\colorbox[HTML]{FF9800}{ff9800}/\colorbox[HTML]{2196f3}{2196f3}/\colorbox[HTML]{ffeb3b}{ffeb3b}/
\colorbox[HTML]{00bcd4}{00bcd4}/\colorbox[HTML]{6f42c1}{6f42c1}/\colorbox[HTML]{e91e63}{e91e63}/\colorbox[HTML]{4caf50}{4caf50} \\
\midrule

2: Text Color &
Text color in the target item &
\#\textcolor[HTML]{6f42c1}{6f42c1}/\textcolor[HTML]{111111}{111111}/\textcolor[HTML]{198754}{198754}/\textcolor[HTML]{dc3545}{dc3545}/\textcolor[HTML]{0d6efd}{0d6efd} \\
\midrule

3: Font Family &
Typeface used for text rendering &
{\fontfamily{pag}\selectfont inter}/{\fontfamily{pag}\selectfont opensans}/
{\fontfamily{phv}\selectfont roboto}/{\fontfamily{phv}\selectfont arial}/
{\fontfamily{phv}\selectfont helvetica}/{\fontfamily{pbk}\selectfont merriweather}/{\fontfamily{pbk}\selectfont georgia}/{\fontfamily{ptm}\selectfont times}/
{\fontfamily{pcr}\selectfont jetbrains-mono}/{\fontfamily{pag}\selectfont verdana}/{\fontfamily{cmss}\selectfont comic}/{\fontfamily{cmss}\selectfont lucida}/{\fontfamily{pcr}\selectfont courier} \\
\midrule

4: Font Size &
Font size of texts in the target item &
14/16/18/20/24px \\
\midrule

5: Position &
Item position on the webpage &
banner/header/sidebar \\
\midrule

6: Card Size &
Scaling factor applied to item layout &
card size scale 0.8/1.2/1.5 \\
\midrule

7: Clarity &
Visual sharpness or blur level &
card\_clarity\_blur\_1/2/4px, image\_clarity\_blur\_1/2/4/8px,
card/image\_clarity\_sharp,
image\_clarity\_very\_sharp \\
\midrule

8: Order &
Order among all the items &
order middle/last \\
\midrule

\end{tabular}
}
\label{tab:variants}
\end{table*}

%% file: contents/4.exp.tex
\section{Experiments}
\subsection{Experiment Setups}

\begin{figure*}[t!]
    \centering
    \includegraphics[width=1.0\linewidth]{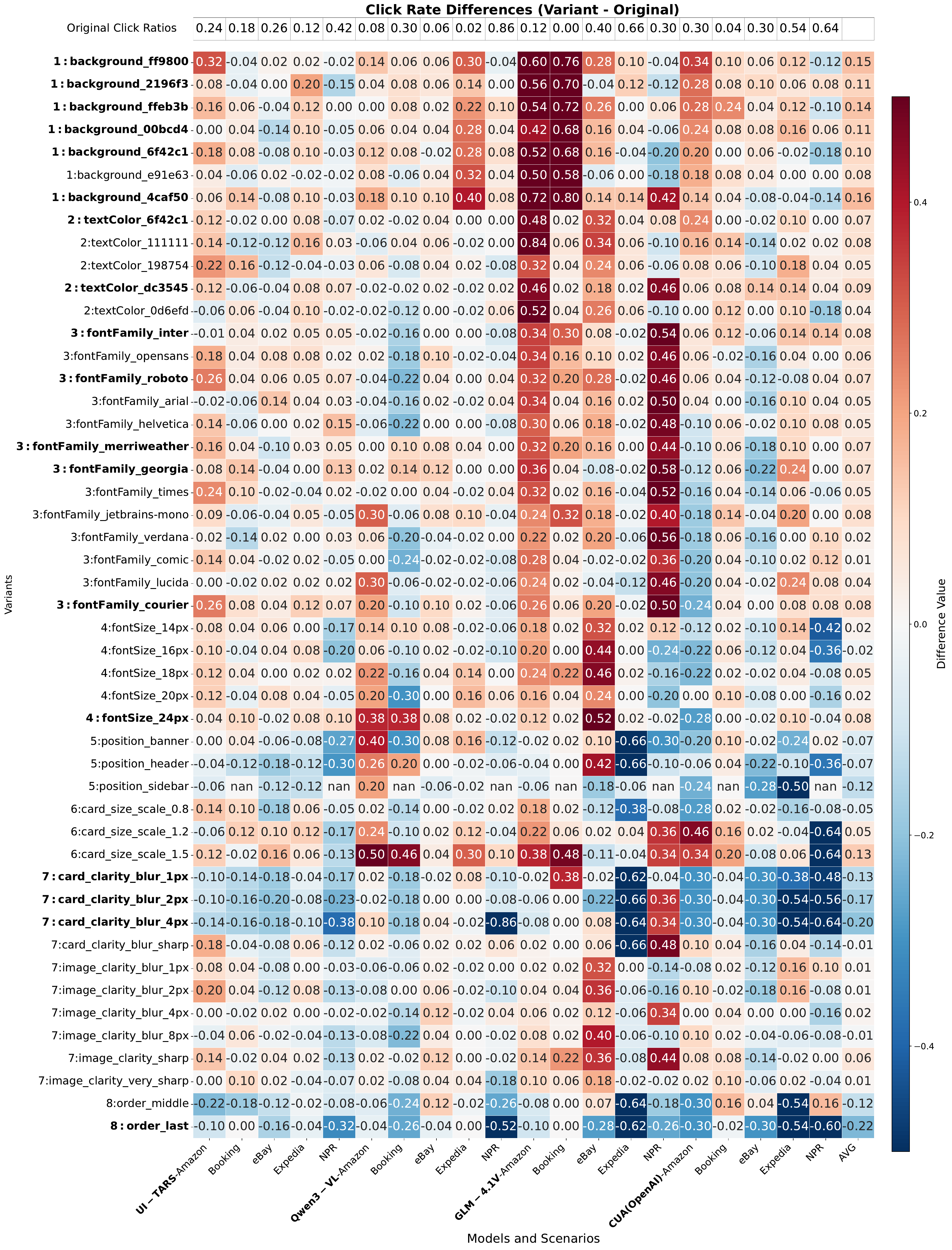}
    \caption{Heatmap of $\Delta=\mathrm{TCR}_{\mathrm{variant}}-\mathrm{TCR}_{\mathrm{original}}$. Larger $\Delta$ indicates that the variant gets more clicks than the original target item. Across diverse models and scenarios, we observe that (1) high background color contrast and enlarged item card consistently increase attraction; (2) item position strongly affects decisions, with agents biased toward selecting the first few items; (3) item image clarity has limited impact, whereas entire card clarity has a stronger effect on agent actions; and (4) font style and text color variants do not have consistent positive/negative effect on decision-making in general. 
    \texttt{nan} indicates variants are not applicable in the corresponding scenario. Bolded variants differ significantly from the original, with $\mathrm{TCR}_{\mathrm{variant}}$ higher/lower than $\mathrm{TCR}_{\mathrm{original}}$ under a one-sided hypothesis test at a 0.05 significance level.
    % the exact p values can refer to Appendix ~\ref{sec:app_Ex}.
    }
    \label{fig:heatmapbig}
\end{figure*}

\noindent\textbf{Real-world websites.} We conduct experiments on five diverse, real-world websites spanning e-commerce (Amazon and eBay), travel (Booking and Expedia), and news browsing (NPR). These websites vary substantially in layout structure and information density, enabling us to study whether the visual attributes consistently impact web agents across different web environments. Details of the webpage and target item are in Appendix~\ref{sec:app_Ex_website}.

\noindent\textbf{Variant Families.} We generate 48 variants across 8 variant families for each target item. Details are shown in Tab.~\ref{tab:variants} and Appendix~\ref{sec:app_Ex_variants}. We only modify the CSS element of the target item, leaving the semantics identical to the original item.

\noindent\textbf{Models.} We select four SOTA VLMs to conduct the experiments: three representative open-source models (UI-TARS 7B~\cite{qin2025ui}, GLM-4.1v-9B~\cite{hong2025glm} Qwen3-VL-8B-Instruct~\cite{bai2025qwen3vltechnicalreport}), and one commercial closed-source model (OpenAI-CUA~\cite{openai_cua_api}). All models are capable of perceiving complex web layouts as visual input and generating reasoning steps followed by actions.

\noindent\textbf{Implementation Details.} We run inference with a temperature of 1.0 and top-$p$ of 0.8 across all the models. The first item displayed on the webpage is selected as the target for variant generation and evaluation. Each variant is evaluated over 50 independent trials. We employ Qwen/Qwen3-14B~\cite{yang2025qwen3technicalreport} as the LLM-as-a-judge, deciding whether web agents notice the target item in their CoT. Complete prompts are in Appendix~\ref{sec:app_Ex_prompt}.

% These tasks reflect common user goals and require multi-step interaction, visual grounding, and semantic understanding, including filtering, scrolling, comparison, and selection. The selected websites vary substantially in layout structure, visual density, and interaction patterns, enabling us to study how different interface designs affect agent perception, decision-making, and robustness in realistic web environments. To control task complexity and interaction cost, we restrict filtering options and limit the number of visible items to fewer than 15, ensuring that agents can make decisions without excessive interactions that may degrade efficiency or performance.

\input{contents/table/top_variant_main}

\subsection{Experimental Results}

\subsubsection{Target Item Click Analysis}

\begin{figure}[t]
    \centering
    \begin{subfigure}[t]{0.48\columnwidth}
        \centering
        \includegraphics[width=\linewidth]{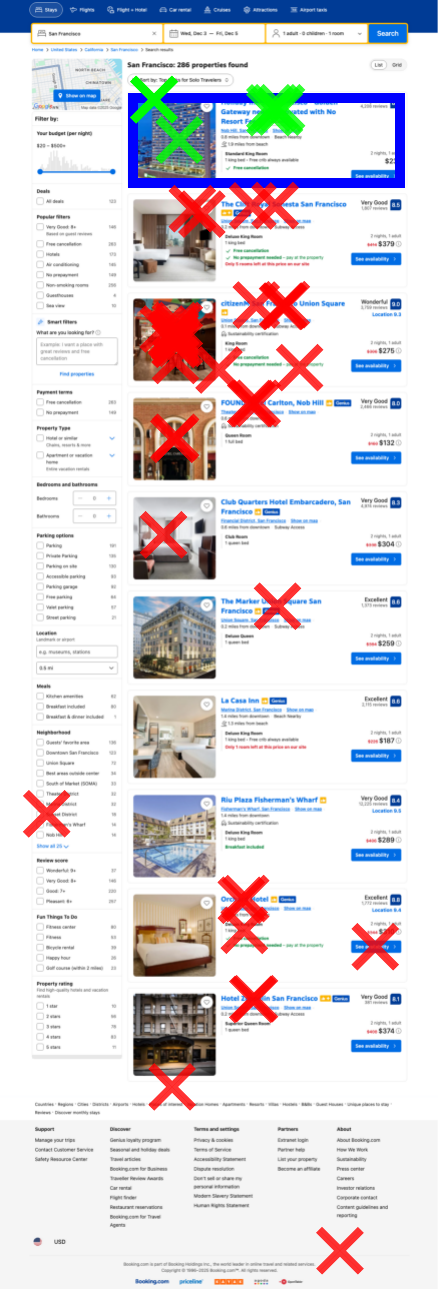}
        \caption{Original Webpage}
        \label{fig:ori}
    \end{subfigure}
    \hfill
    \begin{subfigure}[t]{0.48\columnwidth}
        \centering
        \includegraphics[width=\linewidth]{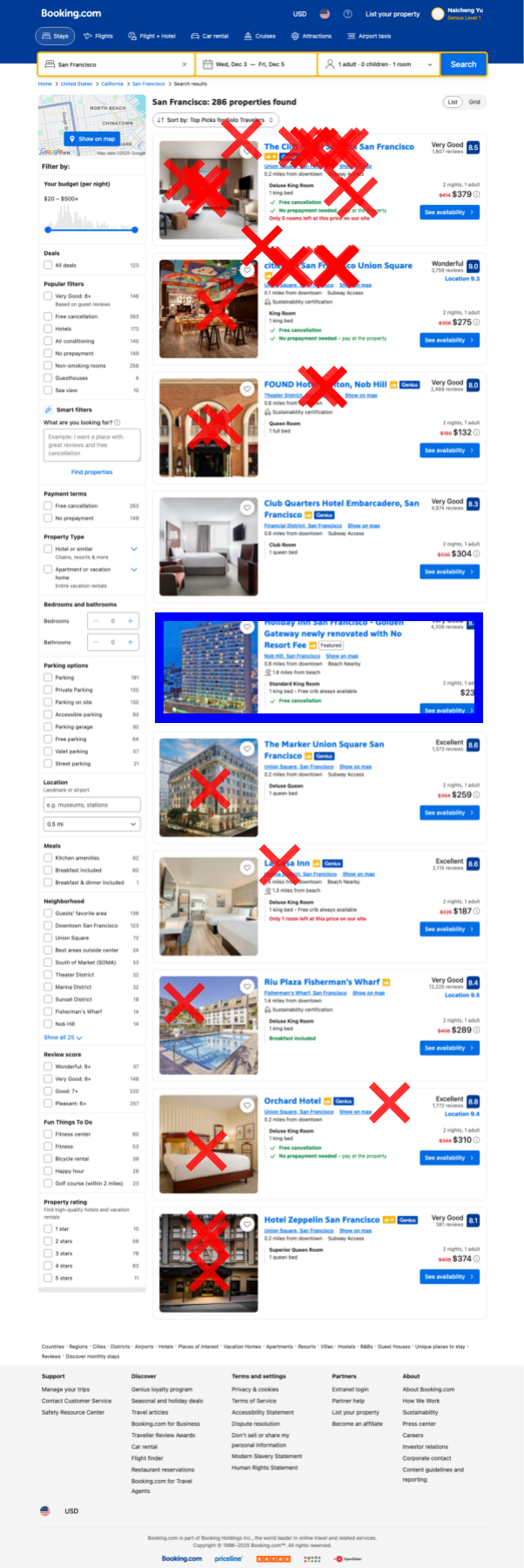}
        \caption{Variant middle\_order}
        \label{fig:middle}
    \end{subfigure}

    \caption{A qualitative example of click distribution comparison on the Booking. The target item (i.e., first item on the original webpage) is marked with a blue frame. After variant generation, the target click rate of UI-TARS 7B decreases from 18\% to 0\% on average across 50 trials, suggesting that the agent tends to click items near the top of the page.}
    \vspace{-15pt}
    \label{fig:discompara}
\end{figure}

As shown in Fig.~\ref{fig:heatmapbig}, we visualize the heatmap of TCR difference between the variant and original item (i.e., $\Delta=\mathrm{TCR}_{\mathrm{variant}}-\mathrm{TCR}_{\mathrm{original}}$). The larger $\Delta$ indicates that the variant attracts the agents and gets more clicks than the original target item. We also provide $\mathrm{TCR}_{\mathrm{original}}$ across diverse models and scenarios at the top of Fig.~\ref{fig:heatmapbig}. Compared to existing research on how visual attributes impact human behavior, we observe that: 

\begin{itemize}
    \vspace{-5pt}
    \item High background color contrast and enlarged item cards consistently increase attraction. In Fig.~\ref{fig:heatmapbig}, the average improvement $\Delta$ reaches 11.7\% over 7 tested background colors, indicating that agents are more likely to click items presented on high-contrast backgrounds. Item size exhibits a similar trend: increasing the card-scale factor from 0.8 to 1.2 and 1.5 raises the $\mathrm{TCR}_{\mathrm{variant}}$ by 12\% and 20\%, respectively, suggesting that agents’ click behavior is sensitive to the visual prominence of the target item. This finding also parallels human visual attention: color contrast and element size are dominant bottom-up cues guiding early visual fixation and click behavior~\cite{soegaard2020visual,wu2003improving,ng2024color}. Human eye-tracking experiments consistently report that high-saturation colors and larger visual footprints attract earlier fixations and long dwell time, thereby increasing the probability of interaction. This echoes the same saliency-driven mechanism of web agents. 
    % However, unlike humans who can suppress saliency when it conflicts with task intent, agents appear more tightly coupled to visual prominence. Once an item becomes visually dominant through color and size, agents disproportionately favor it, even when competing items are semantically plausible. This suggests that agents rely more strongly on bottom-up saliency cues and exhibit weaker top-down control than human users.
    \vspace{-7pt}
    \item  Item position has a strong effect on agent decision: agents are more likely to click items that appear early in the webpage. When we move the target item from the top to the middle or bottom of the page, target click rates consistently decline across nearly all agents and scenarios. Fig.~\ref{fig:discompara} provides a qualitative visualization of click distributions, showing that clicks concentrate on the first few items; correspondingly, shifting the target item to later positions dramatically reduces the probability that it is selected. The strong positional bias observed in agents mirrors classic human browsing behavior, such as F-shaped attention patterns and banner blindness~\cite{pernice2018banner}. Human users allocate most attention to early, central regions of a webpage and systematically under-attend to lower-ranked items, sidebars, or banner-like regions. This indicates that agents and humans have internalized similar layout priors.
    % Our results indicate that agents have internalized similar layout priors. 
    % The key difference lies in recovery behavior. Humans often compensate for positional bias by scrolling, re-evaluating later items, or revisiting ignored regions when early candidates fail. Agents, in contrast, rarely recover once agents fail to notice the item at first attention trial, suggesting a lack of adaptive search strategies and self-correction mechanisms that characterize human goal-directed browsing.
    \vspace{-7pt}
    \item Item image clarity has a limited impact, whereas overall card clarity more strongly affects agent actions. Comparing $\Delta$ for \texttt{card\_clarity\_blur} versus \texttt{image\_clarity\_blur}, we observe that blurring only the item image rarely reduces TCR, while blurring the entire item card substantially decreases TCR in nearly all settings. This suggests that, during browsing, agents rely heavily on the card’s textual content when deciding what to click. This observation diverges from human perception studies showing that moderate image degradation does not necessarily prevent object recognition, as humans can rely on contextual inference~\cite{yoshihara2023blurtraining}. 
    % Similarly, blurring only the item image has limited effect on agent TCR.
    However, when the entire card, including textual content, is blurred, agent performance degrades sharply. This divergence highlights a fundamental difference: humans can compensate for visual degradation through semantic reasoning and prior knowledge, whereas agents depend heavily on clear textual signals to ground their decisions. Once text clarity is compromised, agents struggle to recognize or mention the target at all, indicating that their multimodal understanding remains text-centric and brittle under visual noise.
    \vspace{-7pt}
    \item Font-style and text-color variants generally exert only a minor influence on agent decision-making. While a few specific variants can noticeably affect certain agents, most font-style and text-color changes do not substantially alter whether the agent clicks the target item. Human studies similarly report that, once basic readability constraints are satisfied, variations in font style or text color have limited influence on task-oriented clicking behavior~\cite{krause2022typefaces}. Users adapt quickly to stylistic differences and prioritize semantic relevance over aesthetic variation. The weak effect observed in agents aligns with human trend at the behavioral level. 
    \vspace{-7pt}
    % However, the underlying cause differs. For humans, robustness to typography arises from flexible language comprehension and contextual integration. For agents, reduced sensitivity to font and color likely reflects abstraction in OCR and text-embedding pipelines, which strip away from stylistic nuance. Thus, while the observed click behavior appears similar, agents achieve it through limited perceptual sensitivity rather than adaptive cognitive processing.
\end{itemize}
Taken together, these findings reinforce that web agents share surface-level attentional biases with humans in sensitivity to color, size, and position. However, agents and humans diverge in their ability to recover from distraction or clarity. Agents remain predominantly driven by bottom-up visual saliency, whereas human browsing integrates saliency with semantic verification and strategic exploration. This gap helps explain why certain UI manipulates disproportionately mislead agents while remaining manageable for human users as long as the UI stays reasonable.

% \noindent\textbf{Model-specific visual attribute preference.}

\subsubsection{Target Item Mention Analysis}

To explain why certain variants act as strong attractors or distractors, we use Target Mention Rate (TMR) to indicate whether the agent explicitly notices the target item and mentions it in its CoT. The higher the TMR, the more the agent notices the target item during the interaction. On the contrary, low TMR suggests that the target is overlooked or overshadowed due to other salient items. We provide variants with top/bottom-5 click rate with corresponding TMR in Tab.~\ref{tab:top_bottom_variants_models}.

% The Top/Bottom-5 variants are ranked by click rate, while TMR is used to diagnose the underlying attention and grounding behavior: high TMR implies successful target noticing, whereas low TMR suggests that the target is overlooked or overshadowed by other salient elements.

For UI-TARS 7B, the top-5 variants are dominated by larger font sizes and clear font families, all achieving TMR above the baseline. This indicates that improved typography and readability help the agent visually anchor the target during reasoning. In contrast, position, order, and clarity variants significantly reduce TMR, echoing the observation made in other models as well.

For OpenAI CUA, all Top-5 variants correspond to background-color changes and yield substantially higher TMR than the baseline, suggesting that strong color saliency effectively attracts attention and increases target mentioning. Conversely, Bottom-5 variants, mainly card blur, sidebar, and order changes, cause a sharp drop in TMR, indicating that the model often fails to recognize or reference the target at all.

For Qwen3VL-8B, Top-5 variants combine card size scaling and vivid backgrounds, consistently increasing TMR relative to the baseline. This suggests that larger visual footprints and salient colors improve target noticing and thus downstream click success. However, sidebar position variant again appears in the Bottom-5 with very low TMR, reinforcing that layout displacement suppresses target awareness. Additionally, failures under image blur variants indicate that Qwen3VL is more sensitive to image clarity than other models.

For GLM4.1v-9B, overall TMR is low even for the original interface, implying weaker baseline target grounding. Its Top-5 variants are mostly background-color changes, while Bottom-5 variants are dominated by position-related distractors. This pattern suggests that GLM is particularly vulnerable to attention misallocation: distractors not only reduce click rate but also suppress explicit target mentioning in CoT.

Overall, Tab.~\ref{tab:top_bottom_variants_models} shows that high click-rate variants generally correlate with higher TMR, supporting the interpretation that these variants function as attention attractors. Conversely, many bottom-ranked variants exhibit very low TMR, indicating failure modes where the agent is distracted before grounding the target in its reasoning. 
% This distinction separates two error types: (i) the agent notices the target but acts incorrectly, and (ii) the agent fails earlier by not attending to the target at all.
Tab.~\ref{tab:top_bottom10_variants} provides complete top/bottom-10 ranked variants. We further observe that position variants are the strongest distractors across models, while card clarity is more disruptive than image clarity, highlighting that most agents remain highly sensitive to text clarity, underscoring their reliance on textual information for item understanding.

%% file: contents/table/top_variant_main.tex
\begin{table*}[t]
\centering
\renewcommand{\arraystretch}{1.0}
\caption{Top/Bottom-5 variants ranked by Target Click Rate (TCR) and corresponding Target Mention Rate (TMR). Variants are ranked by TCR averaged over five real-world websites. TMR ($\uparrow$) reflects how frequently the agent explicitly mentions the target item in its CoT. Full Top/Bottom-10 results are reported in Tab.~\ref{tab:top_bottom10_variants}.}

\resizebox{0.8\linewidth}{!}{%
\begin{tabular}{|c|l|c|c|l|c|c|}
\toprule
\multirow{2}{*}{\textbf{Rank}} &
\multicolumn{3}{c|}{\textbf{UI-TARS 7B}} &
\multicolumn{3}{c|}{\textbf{OpenAI CUA}} \\
\cline{2-7}
% \midrule

& \textbf{Variants} & \textbf{TCR} & \textbf{TMR} &
\textbf{Variants} & \textbf{TCR} & \textbf{TMR} \\
\midrule
1 & fontSize\_24px        & 0.409 & 0.580 & background\_ff9800 & 0.496 & 0.668 \\
2 & fontSize\_22px        & 0.376 & 0.664 & background\_00bcd4 & 0.488 & 0.644 \\
3 & fontFamily\_courier   & 0.369 & 0.692 & background\_2196f3 & 0.484 & 0.700 \\
4 & fontFamily\_helvetica & 0.367 & 0.684 & background\_42a5f5 & 0.476 & 0.760 \\
5 & fontFamily\_roboto    & 0.365 & 0.642 & background\_1976d2 & 0.473 & 0.790 \\
\midrule
/ & Original (Baseline) & 0.256 & 0.556 & Original (Baseline) & 0.364 & 0.648 \\
\midrule
44 & image\_clarity\_blur\_4px & 0.188 & 0.556 & fontSize\_16px & 0.255 & 0.570 \\
45 & card\_clarity\_sharp      & 0.168 & 0.528 & image\_clarity\_blur\_1px & 0.233 & 0.588 \\
46 & order\_last               & 0.111 & 0.286 & position\_banner & 0.205 & 0.500 \\
47 & position\_header          & 0.080 & 0.400 & position\_header & 0.190 & 0.400 \\
48 & position\_sidebar         & 0.060 & 0.155 & position\_sidebar & 0.055 & 0.450 \\
\bottomrule
\end{tabular}
}

\vspace{0.2em}

\resizebox{0.8\linewidth}{!}{%
\begin{tabular}{|c|l|c|c|l|c|c|}
\toprule
\multirow{2}{*}{\textbf{Rank}} &
\multicolumn{3}{c|}{\textbf{Qwen3VL-8B}} &
\multicolumn{3}{c|}{\textbf{GLM4.1v-9B}} \\
\cline{2-7}
& \textbf{Variants} & \textbf{TCR} & \textbf{TMR} &
\textbf{Variants} & \textbf{TCR} & \textbf{TMR} \\
\midrule
1 & card\_size\_scale\_1.5 & 0.680 & 0.652 & background\_4caf50 & 0.770 & 0.398 \\
2 & background\_4caf50     & 0.610 & 0.702 & background\_ff9800 & 0.660 & 0.352 \\
3 & background\_6f42c1     & 0.550 & 0.544 & background\_f44336 & 0.660 & 0.494 \\
4 & background\_ffeb3b     & 0.490 & 0.612 & fontFamily\_merriweather & 0.640 & 0.493 \\
5 & fontSize\_24px         & 0.430 & 0.646 & background\_ffeb3b & 0.640 & 0.326 \\
\midrule
/ & Original (Baseline) & 0.320 & 0.550 & Original (Baseline) & 0.280 & 0.414 \\
\midrule
44 & fontFamily\_comic & 0.090 & 0.588 & position\_banner & 0.120 & 0.490 \\
45 & fontSize\_14px    & 0.090 & 0.586 & order\_middle & 0.110 & 0.410 \\
46 & position\_sidebar & 0.080 & 0.253 & position\_spotlight & 0.070 & 0.110 \\
47 & image\_clarity\_blur\_2px & 0.070 & 0.598 & background\_9c27b0 & 0.060 & 0.394 \\
48 & image\_clarity\_blur\_1px & 0.060 & 0.566 & order\_last & 0.030 & 0.327 \\
\bottomrule
\end{tabular}
}

\label{tab:top_bottom_variants_models}
\end{table*}

%% file: contents/6.conclusion.tex
\section{Conclusion}
We introduce \oursabbrv{} to study how visual attributes affect web-agent behavior under real-world websites. Across 48 variants, 5 real-world websites, and 4 agents, we find that background color, item size, position, and card clarity strongly influence agent actions, while font styling, text color, and image clarity have limited impact. These effects mirror human attention patterns driven by visual saliency and positional bias, but agents remain brittle and often fail once visual grounding breaks. We hope our insights will inspire future research.
% Our results highlight both shared perceptual priors and gaps between human and agent web interaction.
% We present the first comprehensive framework for evaluating visual robustness and behavioral bias in web agents under controlled interface variations. By constructing systematic CSS and layout perturbations across real-world scenarios, our approach quantifies both coordinate-level precision and semantic correctness, bridging perceptual grounding with decision reasoning. Through large-scale testing on UI-TARS and GLM 4.1, we reveal consistent attentional patterns shaped by color, size, and positional cues, mirroring human visual biases yet exposing agents' fragility to degraded clarity and typography. This study establishes a unified methodology for probing how design elements influence multimodal agents, laying the foundation for future research on trustworthy, human-aligned web interaction systems.

%% file: contents/limitation.tex
\section{Limitations}
Our study focuses on a representative set of widely used web agents and models rather than an exhaustive coverage of all architectures or scales. Due to the computational cost and complexity of multi-scenario evaluation, we do not include larger-scale agents or a broader range of models in the current experiments. However, the proposed framework is model-agnostic and can be readily extended to additional agents and scales, enabling future expansion into a more comprehensive benchmark.

%% file: contents/ethical_statement.tex
\section{Ethical Statement}
This work does not involve personal data. All experiments were conducted using automated agents in controlled environments. AI assistants were used for language refinement only; all technical contributions and analyses were performed by the authors.

Our study investigates how interface design influences agent behavior with the aim of improving robustness and safety, not enabling manipulation. We do not release the full codebase at this time due to ongoing extensions of the benchmark and evaluation framework, but we plan to make it publicly available in a future release.

%% file: contents/table/css_elements_origianl.tex
\begin{table*}[t]
\centering
\caption{Original CSS elements of each real-world website.}
\label{tab:original_css_elements}
\renewcommand{\arraystretch}{1.0}
\setlength{\tabcolsep}{4pt}

\begin{tabular}{
c | c | c | c | c
}
\toprule
\textbf{Scenario} &
\textbf{Background} &
\textbf{Font Size} &
\textbf{Font Family} &
\textbf{Text Color} \\
\midrule
Amazon  & \texttt{\#00000000} & 18px    & Arial        & \texttt{\#0f1111} \\
Booking & \texttt{\#ffffff}  & 16px    & system-ui   & \texttt{\#1a1a1a} \\
eBay    & \texttt{\#00000000} & 14px    & Market Sans & \texttt{\#191919} \\
Expedia & \texttt{\#ffffff}  & 16.38px & Centra No2  & \texttt{\#191e3b} \\
NPR     & \texttt{\#394f78}  & 28.8px  & NPRSans     & \texttt{\#333333} \\
\bottomrule
\end{tabular}
\end{table*}

%% file: contents/table/top_bottom_10.tex
\begin{table*}[t]
\centering
\renewcommand{\arraystretch}{1.0}
\caption{Top/Bottom-10 variants ranked by Target Click Rate (TCR) and their corresponding Target Mention Rate (TMR). Rankings are based on TCR averaged over five real-world websites, while TMR is reported to indicate how often the target is explicitly mentioned in the CoT ($\uparrow$ means higher TMR).}
\resizebox{0.98\linewidth}{!}{%
\begin{tabular}{|c|l|c|c|c|l|c|c|}
\toprule
\multicolumn{4}{|c|}{\textbf{UI-TARS 7B}} & \multicolumn{4}{c|}{\textbf{Qwen3VL-8B}} \\
\midrule
\textbf{Rank} & \textbf{Variants} & \textbf{TCR} & \textbf{TMR} &
\textbf{Rank} & \textbf{Variants} & \textbf{TCR} & \textbf{TMR} \\
\midrule
1  & fontSize\_24px          & 0.409 & 0.580 & 1  & card\_size\_scale\_1.5      & 0.680 & 0.652 \\
2  & fontSize\_22px          & 0.376 & 0.664 & 2  & background\_4caf50          & 0.610 & 0.702 \\
3  & fontFamily\_courier     & 0.369 & 0.692 & 3  & background\_6f42c1          & 0.550 & 0.544 \\
4  & fontFamily\_helvetica   & 0.367 & 0.684 & 4  & background\_ffeb3b          & 0.490 & 0.612 \\
5  & fontFamily\_roboto      & 0.365 & 0.642 & 5  & fontSize\_24px              & 0.430 & 0.646 \\
6  & background\_1976d2      & 0.363 & 0.604 & 6  & background\_ff9800          & 0.410 & 0.674 \\
7  & fontFamily\_times       & 0.342 & 0.704 & 7  & textColor\_111111           & 0.400 & 0.534 \\
8  & fontFamily\_opensans    & 0.334 & 0.580 & 8  & fontSize\_22px              & 0.390 & 0.528 \\
9  & card\_size\_scale\_0.8  & 0.330 & 0.564 & 9  & fontFamily\_jetbrains-mono  & 0.340 & 0.588 \\
10 & fontFamily\_georgia     & 0.327 & 0.604 & 10 & fontSize\_18px              & 0.330 & 0.648 \\
\midrule
-- & Original (Baseline)   & 0.256 & 0.556 & -- & Original (Baseline) & 0.320 & 0.550 \\
\midrule
39 & image\_clarity\_blur\_4px & 0.188 & 0.556 & 39 & order\_last                    & 0.120 & 0.420 \\
40 & order\_middle                    & 0.188 & 0.427 & 40 & fontFamily\_helvetica   & 0.110 & 0.588 \\
41 & fontSize\_14px            & 0.178 & 0.576 & 41 & fontFamily\_arial       & 0.100 & 0.646 \\
42 & background\_e91e63        & 0.175 & 0.580 & 42 & card\_size\_scale\_0.8  & 0.100 & 0.512 \\
43 & fontFamily\_jetbrains-mono& 0.175 & 0.470 & 43 & image\_clarity\_blur\_4px & 0.100 & 0.588 \\
44 & image\_clarity\_blur\_1px & 0.173 & 0.552 & 44 & fontFamily\_comic       & 0.090 & 0.588 \\
45 & card\_clarity\_sharp      & 0.168 & 0.528 & 45 & fontSize\_14px          & 0.090 & 0.586 \\
46 & order\_last                      & 0.110 & 0.286 & 46 & position\_sidebar               & 0.080 & 0.253 \\
47 & position\_header                 & 0.080  & 0.400 & 47 & image\_clarity\_blur\_2px & 0.070 & 0.598 \\
48 & position\_sidebar                & 0.060  & 0.155 & 48 & image\_clarity\_blur\_1px & 0.060 & 0.566 \\
\midrule
\multicolumn{4}{|c|}{\textbf{GLM4.1v-9B}} & \multicolumn{4}{c|}{\textbf{OpenAI CUA}} \\
\midrule
\textbf{Rank} & \textbf{Variants} & \textbf{TCR} & \textbf{TMR} &
\textbf{Rank} & \textbf{Variants} & \textbf{TCR} & \textbf{TMR} \\
\midrule
1  & background\_4caf50       & 0.770 & 0.398 & 1  & background\_ff9800     & 0.496 & 0.668 \\
2  & background\_ff9800       & 0.660 & 0.352 & 2  & background\_00bcd4     & 0.488 & 0.644 \\
3  & background\_f44336       & 0.660 & 0.494 & 3  & background\_2196f3     & 0.484 & 0.700 \\
4  & fontFamily\_merriweather & 0.640 & 0.493 & 4  & background\_42a5f5     & 0.476 & 0.760 \\
5  & background\_ffeb3b       & 0.640 & 0.326 & 5  & background\_1976d2     & 0.473 & 0.790 \\
6  & fontFamily\_roboto       & 0.630 & 0.415 & 6  & background\_ffeb3b     & 0.470 & 0.636 \\
7  & background\_00bcd4       & 0.610 & 0.491 & 7  & fontSize\_22px         & 0.440 & 0.730 \\
8  & fontFamily\_arial        & 0.560 & 0.433 & 8  & background\_9c27b0     & 0.445 & 0.720 \\
9  & fontSize\_24px           & 0.550 & 0.460 & 9  & textColor\_dc3545      & 0.420 & 0.696 \\
10 & background\_1976d2       & 0.540 & 0.499 & 10 & background\_e91e63     & 0.424 & 0.656 \\
\midrule
-- & Original (Baseline)   & 0.280 & 0.414 & -- & Original (Baseline) & 0.364 & 0.648 \\
\midrule
39 & card\_clarity\_blur\_1px & 0.150 & 0.320 & 39 & fontSize\_16px         & 0.255 & 0.570 \\
40 & position\_header               & 0.140 & 0.920 & 40 & image\_clarity\_blur\_1px & 0.233 & 0.588 \\
41 & card\_clarity\_blur\_4px & 0.130 & 0.102 & 41 & position\_banner               & 0.205 & 0.500 \\
42 & position\_sidebar              & 0.130 & 0.420 & 42 & position\_header               & 0.190 & 0.400 \\
43 & card\_clarity\_blur\_2px & 0.130 & 0.225 & 43 & position\_spotlight            & 0.095 & 0.233 \\
44 & position\_banner               & 0.120 & 0.490 & 44 & position\_sidebar              & 0.055 & 0.450 \\
45 & order\_middle                  & 0.110 & 0.410 & 45 & card\_clarity\_blur\_1px & 0.064 & 0.364 \\
46 & position\_spotlight            & 0.070 & 0.110 & 46 & card\_clarity\_blur\_2px & 0.028 & 0.084 \\
47 & background\_9c27b0      & 0.060 & 0.394 & 47 & card\_clarity\_blur\_4px & 0.020 & 0.064 \\
48 & order\_last                    & 0.030 & 0.327 & 48 & order\_last                     & 0.008 & 0.148 \\
\bottomrule
\end{tabular}%
}
\label{tab:top_bottom10_variants}
\end{table*}